\documentclass[letterpaper, 10 pt, conference]{ieeeconf}  

\IEEEoverridecommandlockouts                              

\usepackage{graphicx} 

\title{\LARGE \bf
The Invisible Map: Visual-Inertial SLAM with Fiducial Markers for Smartphone-based Indoor Navigation
}

\author{Paul Ruvolo$^{1}$, Ayush Chakraborty$^{1}$, Rucha Dave$^{1}$, Richard Li$^{1}$, \\Duncan Mazza$^{1}$, Xierui Shen$^{1}$, Raiyan Siddique$^{1}$, and Krishna Suresh$^{1}$
\thanks{$^{1}$Olin College of Engineering, 1000 Olin Way, Needham, MA USA.}%
}
\usepackage{amssymb}
\usepackage{amsmath}
\usepackage{bm}
\usepackage{cleveref}
\begin{document}
\newcommand{\xt}{\bm{x^{(t)}}}
\newcommand{\xtplusone}{\bm{x^{(t+1)}}}

\maketitle
\thispagestyle{empty}
\pagestyle{empty}

\begin{abstract}

We present a system for creating building-scale, easily navigable 3D maps using mainstream smartphones. In our approach, we formulate the 3D-mapping problem as an instance of Graph SLAM and infer the position of both building landmarks (fiducial markers) and navigable paths through the environment (phone poses). Our results demonstrate the system's ability to create accurate 3D maps. Further, we highlight the importance of careful selection of mapping hyperparameters and provide a novel technique for tuning these hyperparameters to adapt our algorithm to new environments.
\end{abstract}

\section{Introduction}

Indoor 3D mapping and navigation are of vital importance in a number of applications including autonomous mobile robots which need to both accurately determine their position within complex indoor environments and construct optimal plans to move towards a desired destination. Additionally, indoor navigation for pedestrians is becoming increasingly important, e.g., for folks who are navigating in large, unfamiliar environments or for folks who have difficulty navigating (e.g., folks who are blind or low vision).

The task of creating a map of an environment while simultaneously localizing a robot (or other device) within that environment is known as the \emph{Simultaneous Localization and Mapping (SLAM) problem} \cite{pritsker1984introduction}, and many researchers have contributed to the vast body of literature on SLAM. Much of the work has focused on methods that are well-suited to wheeled, mobile robots. In particular, researchers have focused on approaches that fuse wheel-encoder-based odometry with data from range sensors \cite{thrun2002probabilistic}. In our work, we focus on methods that work well for mapping, localization, and navigation with modern smartphones, which can use visual-inertial odometry \cite{huang2019visual} to estimate their own motion as well as identify landmarks using onboard cameras.

Visual-inertial odometry (VIO) is a technique whereby inertial measurements from accelerometers, gyroscopes, and magnetometers are fused with motion constraints derived from image processing to estimate the change in device pose between subsequent video frames. By propagating this estimate from frame to frame, one can compute the device pose relative to its starting location. While VIO algorithms on modern smartphones are known to be quite accurate \cite{kim2022benchmark}, they suffer from the same limitations as odometric approaches to motion estimation: only relative motion is computed requiring additional information to localize within a known environment and small errors in motion estimates accumulate over time to cause large scale localization failures.

\begin{figure}
    \centering
    \includegraphics[height=1in]{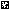}
    \includegraphics[height=1in]{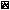}
    \includegraphics[height=1in]{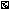}
    \caption{Three of the AprilTags used as fiducial markers by our mapping system. The 6 degree-of-freedom position of these tags can be accurately detected by a smartphone camera across a range of lighting conditions.}
    \label{fig:sampletags}
\end{figure}

To combat the shortcomings of VIO, we use fiducial markers (or ``tags'') with planar patterns that can be readily identified in a camera image. These fiducial markers, called April Tags \cite{wang2016apriltag} \cref{fig:sampletags}, produce best-in-class orientation detection \cite{kalaitzakis2021fiducial} (which is the primary concern for our application). We selected tags based on three factors: (1) there are comparatively fewer fiducial markers-based SLAM approaches (see \Cref{sec:relatedwork}), (2) tags are distinct landmarks for people who are blind or visually impaired navigating unfamiliar indoor environments (3) fiducial markers are well-suited to dynamic indoor environments with highly variable lighting conditions.

\textbf{Contributions:}
\begin{enumerate}
    \item Fiducial-based SLAM system optimized for pedestrian navigation in building-scale indoor environments
    \item Explore the significance of hyperparameter tuning for effective navigation and provide methods to tune hyperparameters with and without a ground truth dataset
    \item Deploy fiducial-based SLAM to a publicly available smartphone app for map creation and navigation.
\end{enumerate}

\section{Related Work}\label{sec:relatedwork}


Tag SLAM \cite{pfrommer2019tagslam} is an approach to fiducial-based SLAM that uses a factor graph to fuse observations of tags at multiple time points and from multiple cameras. In contrast to our approach, Tag SLAM is primarily optimized for the case where multiple tags can be seen at all times by at least one of the cameras. However, we assume sparse tags that are occasionally visible, which is more suited to the case of indoor, pedestrian navigation.

The work of Wang et al. \cite{wang2023robust} builds on Tag SLAM by better handling pose ambiguities in the detection of fiducial markers. In contrast to our work, Wang et al.'s method requires tags to be detected on each frame, and if tags are not detected a relocalization process is needed.

Uco Slam \cite{munoz2020ucoslam} is an approach to combining keypoint-based mapping with fiducial markers. The system shows good performance for sequences both with and without fiducial markers. \cite{munoz2020ucoslam} uses tag corner reprojection errors which we find to be less robust when high odometry drift occurs between loop closures. Further, we additionally explore the challenge of hyperparameter selection and the impact on map quality.

\section{Methods}\label{sec:methods}
In this section, we describe the processes for map generation, user localization, and shortest-path construction for navigation to a desired location. The core of our approach is based on Graph SLAM, specifically the g2o algorithm \cite{grisetti2011g2o}. We use Graph SLAM as a means of combining odometry data with observations of fiducial markers (April Tags \cite{wang2016apriltag}).

\subsection{Map Generation}
During the mapping process, the user walks through the environment and captures the location of fiducial markers by detecting the markers with the phone's camera (using the VISP3 library \cite{marchand2005visp} for tag detection and pose estimation). Additionally, the user can indicate points of interest within the environment (e.g., particular rooms in the building or other building amenities) by placing the phone in the appropriate location and entering information via an on-screen text box. The data collected from this process is fed into our mapping algorithm, which creates a map consisting of the time series of 3D device poses and the 3D landmark poses.

\subsubsection{Graph SLAM Formulation}

Graph SLAM approaches typically consist of two components: a frontend, which translates available sensor data to specific constraints on the robot motion and the poses of landmarks in the environment, and a backend, which optimizes the constraints generated by the frontend to find the best possible map.

\paragraph{Notation}

We denote the 6-degree-of-freedom pose of the phone (or robot) as $\xt$. Specifically, each $\xt \in \mathbb{R}^6$ where $\xt = \left [p^{(t)}_x, p^{(t)}_y, p^{(t)}_z, q^{(t)}_x, q^{(t)}_y, q^{(t)}_z \right]^\top$. $\bm{p^{(t)}} = \left [p^{(t)}_x, p^{(t)}_y, p^{(t)}_z\right]^\top$ represents the 3d-position at time $t$ and $\bm{q^{(t)}} = \left [q^{(t)}_x, q^{(t)}_y, q^{(t)}_z\right]^\top$ is a compact quaternion representing the 3d-orientation. The pose of the $i$th landmark is represented as $\bm{y^{(i)}} \in \mathbb{R}^6$. We use the notation $\mathcal{X}$ to refer to the set of all landmark and phone poses, which form the optimization variables in our SLAM formulation.

Our frontend constraints are quadratic functions whose coefficients are given by importance matrices. We write these importance matrices as $\bm{\Lambda}$ and we use the notation $\bm{\Lambda}(\Theta)$ to make explicit the dependence of these importance matrices on algorithm hyperparameters $\Theta$.

The reference measurements (e.g., those obtained from the phone's odometry), which provide the soft constraints on the relative pose of consecutive phone poses are written as $\bm{m}^{(t,t+1)} \in \mathbb{R}^6$ (the measured relative pose of the phone at time $t+1$ expressed in terms of the measured pose of the phone at time $t$).

\paragraph{SLAM Frontend}\label{sec:frontend}
Our SLAM frontend includes three types of constraints that define our optimization function.

\begin{align}
    \ell(\mathcal{X}) &= \ell_{odom}(\mathcal{X}) + \ell_{gravity}(\mathcal{X}) + \ell_{tag}(\mathcal{X})
\end{align}

The function $\ell_{odom}(\mathcal{X})$ encodes soft constraints generated by our odometry measurements and is defined as:
\begin{align}
  \ell_{odom}(\mathcal{X}) &= \sum_{t=1}^{T-1} \bm{e}_{odom}^{(t, t+1)\top} \bm{\Lambda}_{odom}  \bm{e}_{odom}^{(t, t+1)} \enspace . \\
  \bm{e}_{odom}^{(t, t+1)} &= \bm{m}^{(t, t+1)} - \bm{r}(\xt, \xtplusone) \nonumber
\end{align}

Where $\bm{r}(\xt, \xtplusone)$ represents the relative position.

$\ell_{tag}(\mathcal{X})$ encodes soft constraints generated by our tag measurements and is defined as:

\begin{align}
  \ell_{tag}(\mathcal{X}) &= \sum_{i=1}^{N} \bm{e}_{tag}^{(i)\top} \bm{\Lambda}_{tag}  \bm{e}_{tag}^{(i)} \enspace . \\
  \bm{e}_{tag}^{(i)} &= \bm{m}^{(i)} - \bm{r}(\bm{x}^{(a_i)}, \bm{y}^{(b_i)}) \nonumber
\end{align}
For the $i$th tag measurement $a_i$ denotes the index of the phone pose and $b_i$ denotes the index of the landmark.

Finally, the gravity vector in the phone frame can be observed drift-free using the phone's accelerometer which we add as an additional loss function $\ell_{gravity}(\mathcal{X})$:

\begin{align}
    \ell_{gravity} &= \sum_{t=1}^T e_{gravity}^{(t)\top} \bm{\Lambda}_{gravity} e_{gravity}^{(t)} \enspace .\\
    \bm{e}_{gravity}^{(t)}  &= \bm{m}_{gravity}^{(t)} - \bm{f}_{gravity}(\xt) \nonumber
\end{align}

Where $\bm{m}_{gravity}^{(t)}$ represents the measurement of the gravity vector in the phone's frame at time $t$ and $\bm{f}_{gravity}(\xt)$ represents the predicted measurement based on a candidate phone pose, $\xt$.

\begin{figure}
    \centering
    \includegraphics[width=1.0\linewidth]{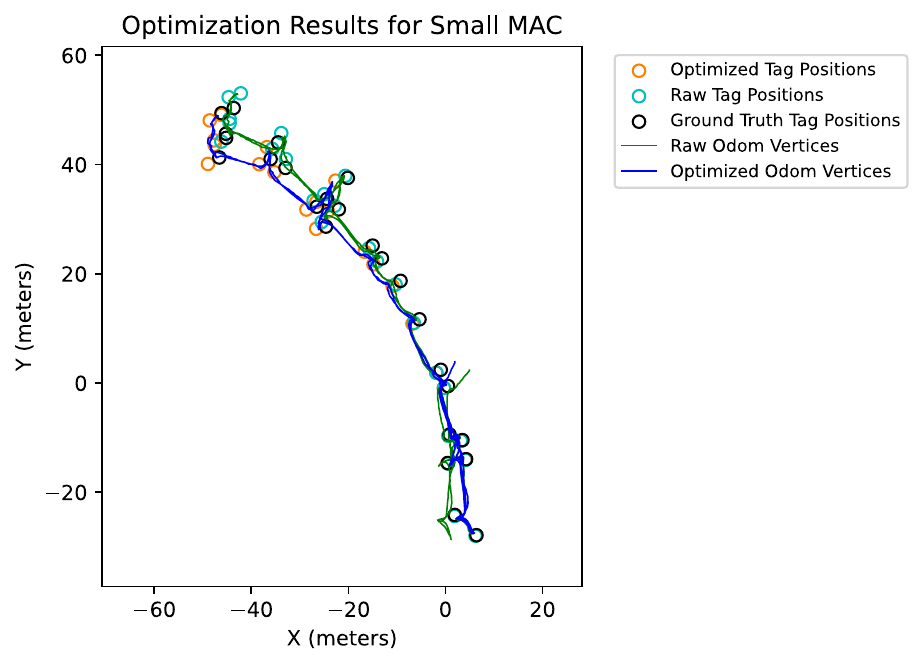}
    \caption{Visualization of Backend Optimization on Small MAC Dataset. This visualization is indicative of how the optimization shifts tag and odometry positions to more acccurately reflect the physical environment.}
    \label{fig:optimization_vis}
\end{figure}

\paragraph{SLAM Backend}
For our SLAM backend we use a sparse, block-wise optimizer using Levenberg-Marquardt method \cite{ranganathan2004levenberg} to progressively reduce the map error from the g2o package \cite{grisetti2011g2o}. We specify a set of ``weights'', $\Theta$, which capture our importance given to each of the loss objectives. We simplify our weight space by tying various parameter values together to arrive at the following four hyperparameters: tag position variance, tag orientation variance, linear odometry variance, and angular odometry variance. A typical backend optimization result is shown in Figure~\ref{fig:optimization_vis}.

\subsection{Metrics and Map Evaluation Procedures}\label{sec:metrics}

\subsubsection{Ground Truth}

We wanted a quantitative metric of how well maps generated from our system matched the true geometry of the location. This goal necessitated a \emph{ground truth} measurement to quantify this difference. While many approaches exist for measuring ground truth tag positions including motion capture systems or laser-based surveying equipment, we decided to use a higher-accuracy SLAM system (with access to better sensors) as a low-cost alternative. 

Labbé's RTABMap \cite{labbe2019rtab} is a system that fuses LIDAR, VIO, and measurements of fiducial markers to create a 3D map of an environment. \emph{Ground truth metric} (or GT) will be a measure of how much a map generated through our system differs from the map generated by RTABMap. To compute this metric, we iterate over each tag and use it as an ``anchor tag'' to align its pose to its ground truth pose. With both maps in the same coordinate system, we compute the Euclidean difference between all corresponding tags in the two maps. The average normalized translational error is computed with each tag being the anchor and then averaged together to yield the ground truth metric. This can be interpreted as the average error of a tag's position (in meters) for every meter walked in the environment from the last tag observation.

\subsubsection{Shift Metric}

\begin{figure}
\begin{center}  \includegraphics[width=0.48\linewidth]{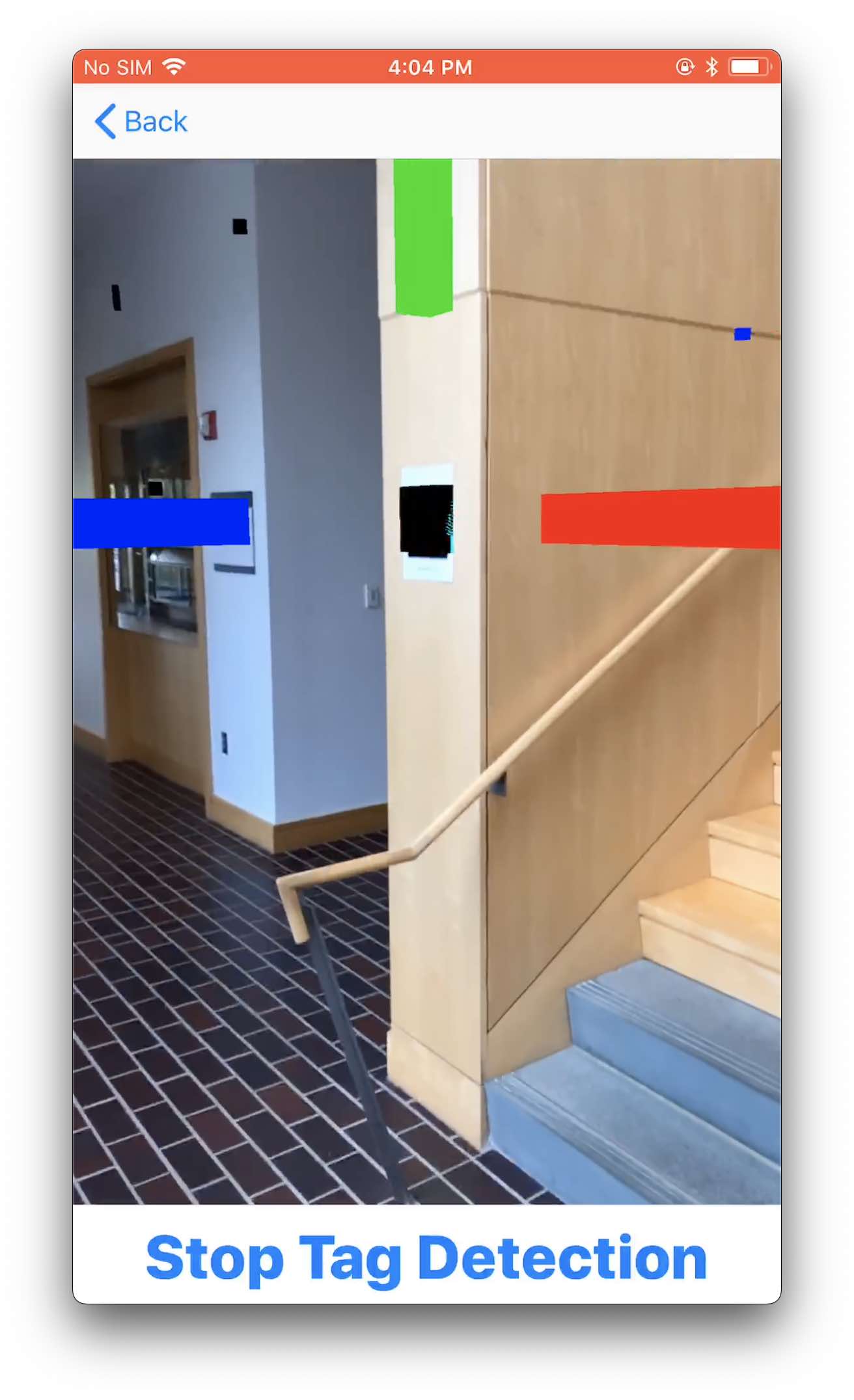}
\includegraphics[width=0.48\linewidth]{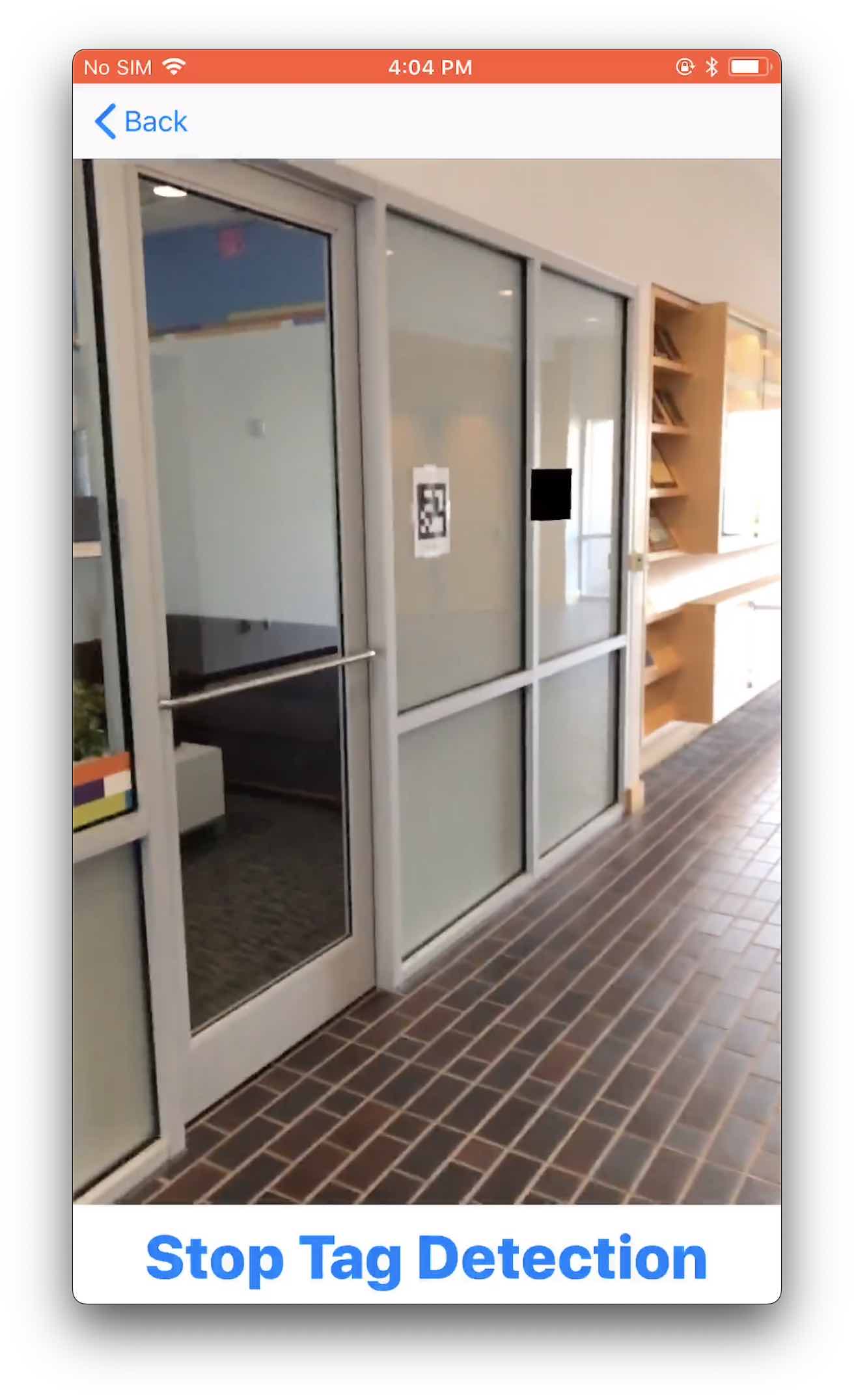}
\end{center}
\caption{Two screenshots from our navigation app that demonstrate the idea of the shift metric. \textbf{Left:} a tag is detected and used to align the map (consisting of tag poses) within the current tracking session. \textbf{Right:} As we approach a second tag, we see that the predicted position of the tag (black square) does not match the physical position of the fiducial marker. The intuition is that better maps will, on average, require smaller shifts to adjust the map when compared to worse maps.}
\label{fig:shiftmetricscreenshots}
\end{figure}

To evaluate and tune our system in novel environments (without access to a LIDAR-enabled iPhone), we developed the shift metric. The shift metric is based on the intuition that with a perfect map, subsequent observations between two tags should result in near-matching relative transformations. While some discrepancy is expected due to imperfect odometry, we would expect the magnitude of the discrepancy to be less for more accurate maps than for less accurate maps. This intuition is captured in Figure~\ref{fig:shiftmetricscreenshots}.

We formalize the intuition above by first anchoring the first tag detections in the unoptimized and optimized map. Then we measure the second detected tag's translational difference and normalize based on the distance from the first tag. This error is calculated for and averaged across all tags, yielding the shift metric for a dataset. This can be interpreted as the average translational error between neighboring tags for every meter apart they are positioned. 

\subsection{Map Navigation}

Our mapping procedure generates the 3D-poses of each April Tag along with the time series of 3D-poses of the phone during the map generation process. In order to navigate within the map, we must address two key problems: localization and path planning.

\subsubsection{Localization}

We formulate the localization problem using a similar graph optimization formulation that we use for map generation. We construct odometric constraints to specify the relative motion of the phone over time. In contrast to our map generation formulation, we consider the tag poses to be fixed. We run the graph SLAM optimizer after each set of tag observations in order to update the estimate of the user's (the phone's) current position. The results of the previous optimization are used as an initial guess for the next optimization (warm start) in order to speed up convergence.

\subsubsection{Path Planning}

Our core assumption in planning is that the path walked by the user during the mapping process represents a navigable path. With two consecutive phone poses generated from our graph SLAM algorithm, we can define a navigable line segment within the map. We represent the collection of line segments as a graph where poses collected subsequently in time are connected via a Euclidean-distance-weighted edge. Additionally, we detect self-intersects and insert additional graph nodes that serve as junctions to connect multiple path segments collected at disparate points in time. The resultant weighted graph can be used to generate the shortest paths from the user's current position to a desired destination within the map.

\section{Experiments}\label{sec:experiments}
To better understand the performance of our system under known conditions, we tested using simulated data and validated with real data collected from a smartphone moving about in a large multi-story building. Each experiment was conducted on a series of four datasets, one large dataset and one small dataset in two different locations Table~\ref{tab:map_desc}.

\begin{table}[!htb]
\caption{Physical descriptions of the four maps.} \label{tab:map_desc}
\setlength\tabcolsep{0pt} 
\footnotesize\centering
\smallskip 
\begin{tabular*}{\columnwidth}{@{\extracolsep{\fill}}rcccr}
      Map Name & Distance Traveled (m) & Floors Traveled & Tag Detections\\
      \hline
      Small WH & 126.56 & 1 & 7\\
      Big WH & 271.97 & 1 & 19\\
      Small MAC & 379.12 & 1 & 25\\
      Big MAC & 1400.3 & 2 & 52\\
\end{tabular*}

\end{table}

\subsection{Simulated Data}

To understand the correlation between our system's performance and noise present in data, we created simulated data by using an actual trajectory of device poses collected from the phone and adding varying degrees of noise. Noisy estimates of tag positions are also generated along the path given some hypothetical locations of tags and we then evaluate based on recovery of the true tag positions using the metrics discussed in Section~\ref{sec:metrics}. 

We first used this simulated environment to evaluate the effectiveness of the shift metric in supplementing the true ground truth metric in hyperparameter selection. A specific trial with artificial noise displayed an equivalent normalized error of $0.01453$ with both the ground truth metric and the shift metric suggesting its applicability as a surrogate in hyperparameter optimization. We observed this same result in simulated scenarios with tag observation noise, however, the shift metric did occasionally yield marginally worse performance in datasets with low noise.

\subsection{Real Data}

To validate our system within physical environments, we first assessed whether there would be a significant improvement in ground truth accuracy through optimization. Next, we hypothesized that the shift metric would provide an effective method for hyperparameter optimization and allow us to adapt our algorithm to novel environments.

\subsubsection{Improvement in Ground Truth Accuracy}

We tested the hypothesis that our system would improve performance with respect to the ground truth metric when compared to a system that simply used the first observation of each unique tag to create the map (we call this approach \emph{Pre-Optimized}). The optimization results of the aforementioned datasets are shown in Table~\ref{tab:pre_opt_versus_post_opt}. These results show a substantial improvement in ground truth accuracy across all four tested datasets.

\begin{table}[!htb]
\caption{optimization results across four datasets, comparing ground truth metrics from pre-optimized, optimized, and shift} \label{tab:pre_opt_versus_post_opt}
\setlength\tabcolsep{0pt} 
\footnotesize\centering

\smallskip 
\begin{tabular*}{\columnwidth}{@{\extracolsep{\fill}}rccccr}
      Map Name & Pre-optimized & Optimized  & Shift\\
      \hline
      Small WH & 0.07507 & \textbf{0.05882}& 0.05953\\
      Big WH & 0.05689 & \textbf{0.04896} & 0.05012\\
      Small MAC & 0.05105 & \textbf{0.03449} & 0.05029\\
      Big MAC & 0.05059 & \textbf{0.03364} & 0.04984\\
\end{tabular*}

\end{table}

\subsubsection{Hyperparameter Selection}

\begin{figure}
    \centering
    \includegraphics[width=\linewidth]{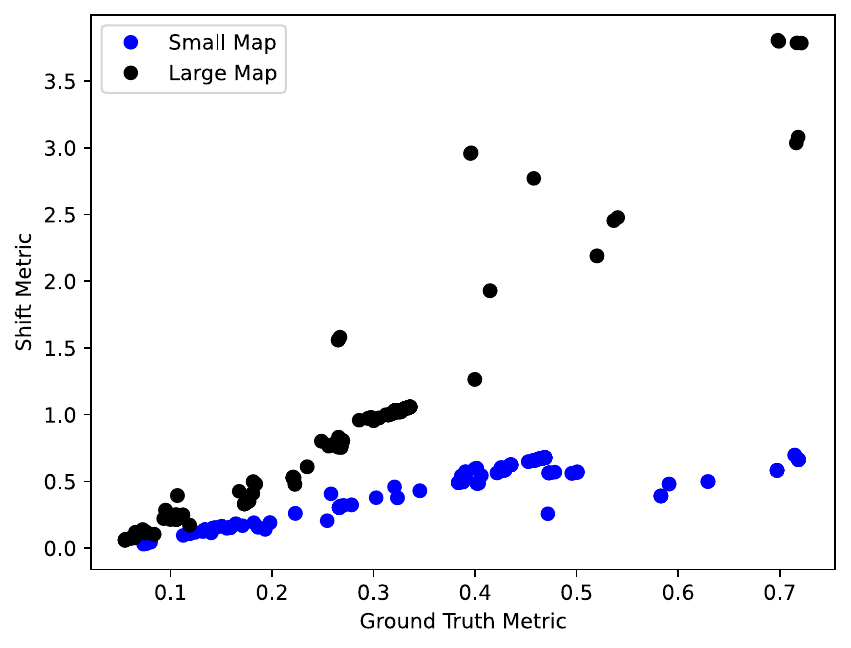}
    \caption{Full sweep results of large and small maps in WH, visualizing a strong positive correlation between the shift metric and ground truth metric. Each point corresponds to a different set of optimization hyperparameters.}
    \label{fig:shift_gt}
\end{figure}

We replicated the experiment in Section~\ref{sec:metrics} with physical data to assess whether the shift metric could serve as a reliable surrogate to the ground truth metric for tuning hyperparameters. Figure~\ref{fig:shift_gt} demonstrates a strong positive correlation between the two performance metrics and suggests the shift metric can be used as a surrogate when choosing hyperparameters.

For each of our four datasets, we computed the ground truth metric with respect to those that directly minimized the ground truth error and those that minimized the shift metric. As shown in Table~\ref{tab:pre_opt_versus_post_opt}, the accuracy obtained when optimizing with the shift metric is close to that obtained when directly minimizing the ground truth metric.




\section{Discussion and Possible Extensions}\label{sec:discussion}

The results shown in Section~\ref{sec:experiments} highlight both the need for careful hyperparameter selection for SLAM with sparse landmarks (e.g., tags placed every 10-20 meters as was done in our experiments) as well as the potential for the shift metric to serve as an effective method for tuning hyperparameters. The shift metric and its ability to tune models to novel environments provide a potential method for adapting our SLAM method based on lighting conditions, camera quality, and other factors specific to a mapping session. We intend to extend this system with hyperparameters selected from easily measurable environmental features and integration of \emph{Google Cloud Anchors} for additional stationary visual features.

\section{Conclusion}

The Invisible Map project is a full-featured indoor mapping and navigation system designed to work with modern smartphones. The system prioritizes robustness at each stage with the eventual goal of providing accessible indoor, turn-by-turn pedestrian navigation. Our system can be deployed easily by users regardless of their technical expertise, and we are currently working to evaluate its usability as an indoor navigation tool for folks who are blind or low vision. More generally, the system can be used in robotics that combine visual-inertial navigation with fiducial markers.




\section{Acknowledgement}

This material is based upon work supported by the National Science Foundation under Grant No. 2007824. We thank William Derksen for contributions to this work.

\bibliographystyle{IEEEtran}
\bibliography{references}

\end{document}